\documentclass[10pt,twocolumn,letterpaper]{article}

\usepackage{cvpr}      

\usepackage{multirow}
\usepackage{booktabs}

\usepackage[dvipsnames]{xcolor}

\definecolor{cvprblue}{rgb}{0.21,0.49,0.74}
\usepackage[pagebackref,breaklinks,colorlinks,citecolor=cvprblue]{hyperref}

\def\paperID{*****} 
\def\confName{CVPR}
\def\confYear{2024}

\title{Dance Your Latents: Consistent Dance Generation through
Spatial-temporal Subspace Attention Guided by Motion Flow}

\author{Haipeng Fang$^{1,2}$,
    Zhihao Sun$^{1,2}$,
    Ziyao Huang$^{1,2}$,
    Fan Tang$^{1,2}$,
    Juan Cao$^{1,2}$,
    Sheng Tang$^{1,2*}$\\
$^{1}$Institute of Computing Technology, Chinese Academy of Sciences\\
$^{2}$University of Chinese Academy of Sciences\\
}

\begin{document}
\maketitle
\begin{abstract}

The advancement of generative AI has extended to the realm of Human Dance Generation, demonstrating superior generative capacities. However, current methods still exhibit deficiencies in achieving spatiotemporal consistency, resulting in artifacts like ghosting, flickering, and incoherent motions. In this paper, we present Dance-Your-Latents, a framework that makes latents dance coherently following motion flow to generate consistent dance videos. Firstly, considering that each constituent element moves within a confined space, we introduce spatial-temporal subspace-attention blocks that decompose the global space into a combination of regular subspaces and efficiently models the spatiotemporal consistency within these subspaces. This module enables each patch pay attention to adjacent areas, mitigating the excessive dispersion of long-range attention. Furthermore, observing that body part's movement is guided by pose control, we design motion flow guided subspace align $\&$ restore. This method enables the attention to be computed on the irregular subspace along the motion flow. Experimental results in TikTok dataset demonstrate that our approach significantly enhances spatiotemporal consistency of the generated videos.

\end{abstract}    
\section{Introduction}
\label{sec:intro}

Recently, diffusion-based generative models \cite{DDPM, Score-based} have garnered considerable attention due to their outstanding performance in image generation \cite{DALLE, SD, Photorealistic}. However, when applied to video generation \cite{vdm, LVDM, ImagenVideo, MakeAVideo, MagicVideo, ControlVideo, VideoLDM, VideoComposer}, and more specifically, Human Dance Generation, prevailing state-of-the-art methods like DreamPose \cite{DreamPose} and DisCo \cite{DisCo} still rely on a frame-by-frame generation approach. Consequently, these approaches often fall short in modeling spatiotemporal consistency, giving rise to artifacts including ghosting, flickering, and incoherent motions.

Several approaches have been proposed to address spatiotemporal consistency. Some methods \cite{TuneAVideo, T2VZero, Pix2Video, FateZero} extend the self-attention module to incorporate multiple frames; however, while this enhancement improves temporal consistency, it slightly compromises the quality of single-frame image generation. Additionally, other methods \cite{MakeAVideo, VideoFusion} expand the 2D U-Net to Pseudo 3D U-Net, a modification that involves swapping the spatial and temporal dimensions to facilitate patch interactions across the identical spatial positions at different temporal positions. Although this strategy effectively enhances the consistency, it struggles with moving objects and sometimes produces visual artifacts such as ghosting and flickering.

\begin{figure}[t] 
	\includegraphics[width=8.4cm]{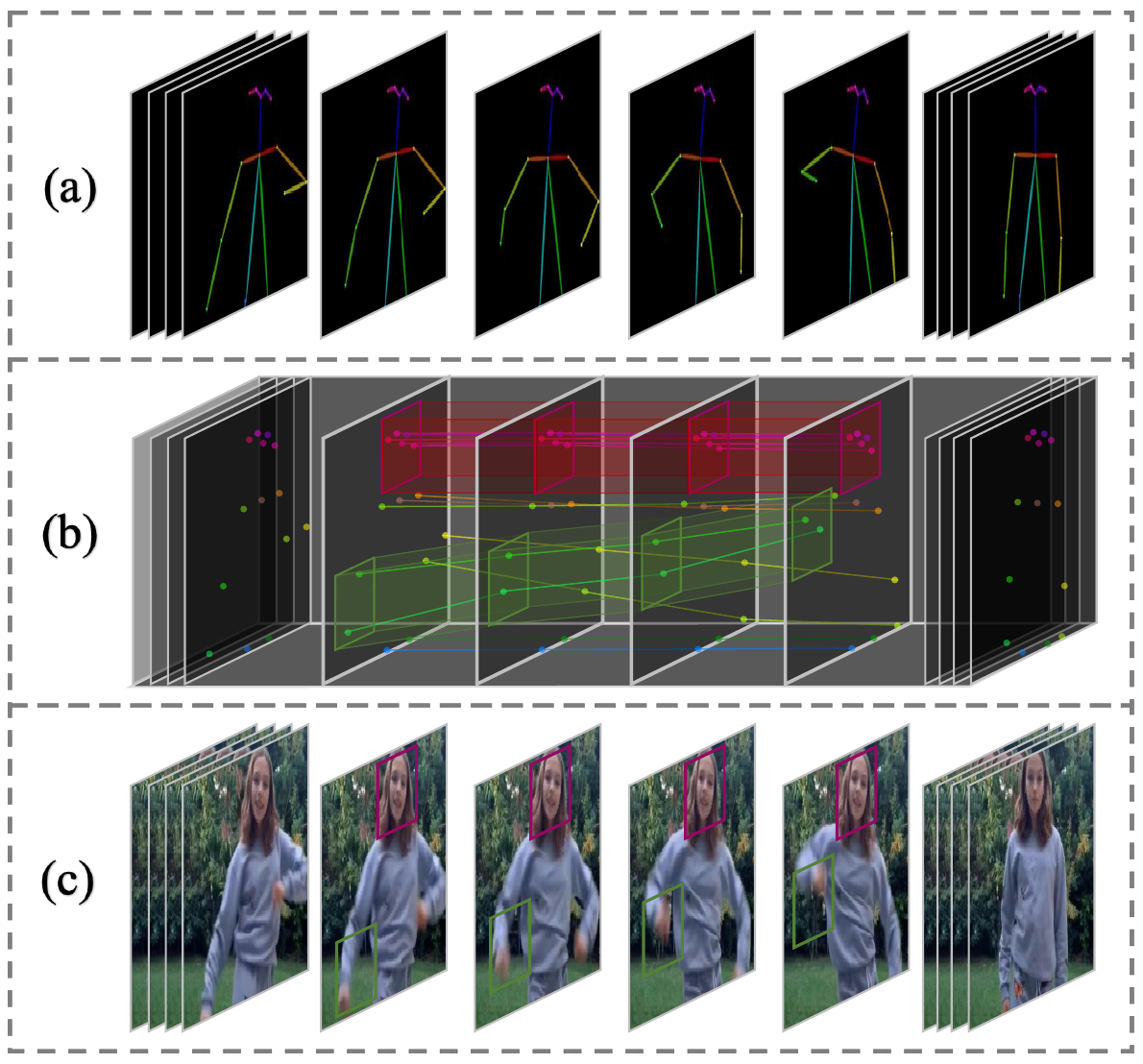}
	\caption{Motivation of Dance-Your-Latents. (a) pose sequence. (b) motion flow. (c) generated video. The black cuboid symbolizes the global space, while the irregular red and green boxes represent subspaces with smaller and larger movements, respectively.}
	\label{fig:motivation}
\end{figure}

In our task, given an image of a human foreground, a background and a sequence of poses, the objective is to synthesize a realistic video, as shown in Fig. \ref{fig:motivation}(a) and Fig. \ref{fig:motivation}(c). The motion flow of each keypoint is illustrated in Fig. \ref{fig:motivation}(b). 

First, we observed that the motion amplitudes of background and the majority of foreground body parts are relatively small, as indicated by the red box in Fig.\ref{fig:motivation}. Besides, some limbs, as illustrated by the green box in Fig.\ref{fig:motivation}, display larger motion amplitudes. This observation leads to the notion that a significant fraction of the video's constituent elements moves within a confined spatiotemporal space. Although a smaller fraction of the elements moves in a larger space, this subspace remains considerably more confined compared to the whole video space. Therefore, we introduce a Spatial-Temporal Subspace-Attention block, abbreviated as STSA. This block aims to simplify the complex task of achieving full-space spatiotemporal consistency by decompose the global space to a combination of subspaces and model the spatiotemporal consistency within these spaces. Additionally, we shift subspaces to propagate the subspaces' consistency across the global space.

Furthermore, considering that video's constituent elements moves in a irregular subspace following the motion flow, as depicted in the green boxes in Fig.\ref{fig:motivation} (b). We aim to transform the irregular subspace to the regular subspace for efficient attention calculation. Therefore, we extract the motion flow from pose sequence, avoiding the interference from the origin video. At the start of STSA, we introduce a Subspace Alignment operation that employs the motion flow to align the content and position within each subspace, and at the end of this STSA module, a Subspace Restoration operation is utilized to recover to the original structure. Our contributions are summarized as follows:
\begin{itemize}
\item We introduce a novel spatial-temporal subspace-attention mechanism. It decomposes the global space into a combination of subspaces, modeling and propagating the consistency within and across these subspaces.
\item We present a motion flow guided subspace alignment and restoration strategy. It aligns these subspaces guided by motion flow extracted from the pose sequence, enabling the efficient attention calculation in irregular subspace.
\item Experiments show that our approach greatly improves spatiotemporal consistency of the generated videos.
\end{itemize}

\section{Related Work}

\subsection{Diffusion Models.} Image and video generation is a basic task in computer vision. Early research employed a variety of generative models, such as Autoregressive Models \cite{AG2, AG1}, Variational Autoencoders \cite{VAE1,VAE2}, Generative Adversarial Networks \cite{GAN1,GAN2,GAN3}, and Normalizing Flows \cite{NF1,NF2}. Diffusion Models \cite{DDPM, Score-based} have shown significant performance in image generation. Notably, Stable Diffusion \cite{SD} leverages latent space diffusion and denoising, concurrently achieving high efficiency and competitive quality. To enhance content control, ControlNet \cite{ControlNet} and T2I-Adapter \cite{T2I-Adapter} incorporate control information, such as skeletons, sketches, and segmentation maps, facilitating precise content control. Moreover, DreamBooth \cite{DreamBooth} empowers users to manipulate subject appearance. Now, diffusion-based models have been further extended into video generation. For instance, Make-A-Video \cite{MakeAVideo} and VideoLDM \cite{VideoLDM} integrate temporal blocks into text-to-image diffusion to facilitate video synthesis. Gen-1 \cite{GEN1} endeavors to transfer video style through a similar temporal architecture. Tune-A-Video \cite{TuneAVideo} involves finetuning on source video while editing prompts to generate the target video. Text2Video-Zero \cite{T2VZero} and FateZero \cite{FateZero} extend text-to-image diffusion with temporal layers, enabling video synthesis without additional training.

\subsection{Human Dance Generation.}

In the context of Human Dance Generation, the objective is synthesizing a realistic video from an individual's image and a sequence of skeletons. Previous studies mainly focus on tasks like motion transfer \cite{MT1,MT2,MT3,MT4,MT5} or still image animation \cite{IA1, IA2, IA3, IA4, IA5}. However, due to the limitations of traditional generative models, cascaded training stages are necessary to decouple the generation of background, motion, and occluded regions of characters in order to ensure effective synthesis. In diffusion-based methods, DreamPose \cite{DreamPose} proposes an appearance-and-pose conditioned diffusion method for animating still fashion images. DisCo \cite{DisCo} disentangles control over appearance, pose, and background to enhance faithfulness and compositionality in dance synthesis. These methods demonstrate the capability to produce high-quality human dance images with diverse appearances and movements. However, these diffusion-based methods generate videos frame-by-frame, leading to unsatisfied consistency and sometimes produce artifacts like ghosting, flickering, and incoherent motions.

\begin{figure*}[!t]
	\centering
	\includegraphics[width=0.95\linewidth]{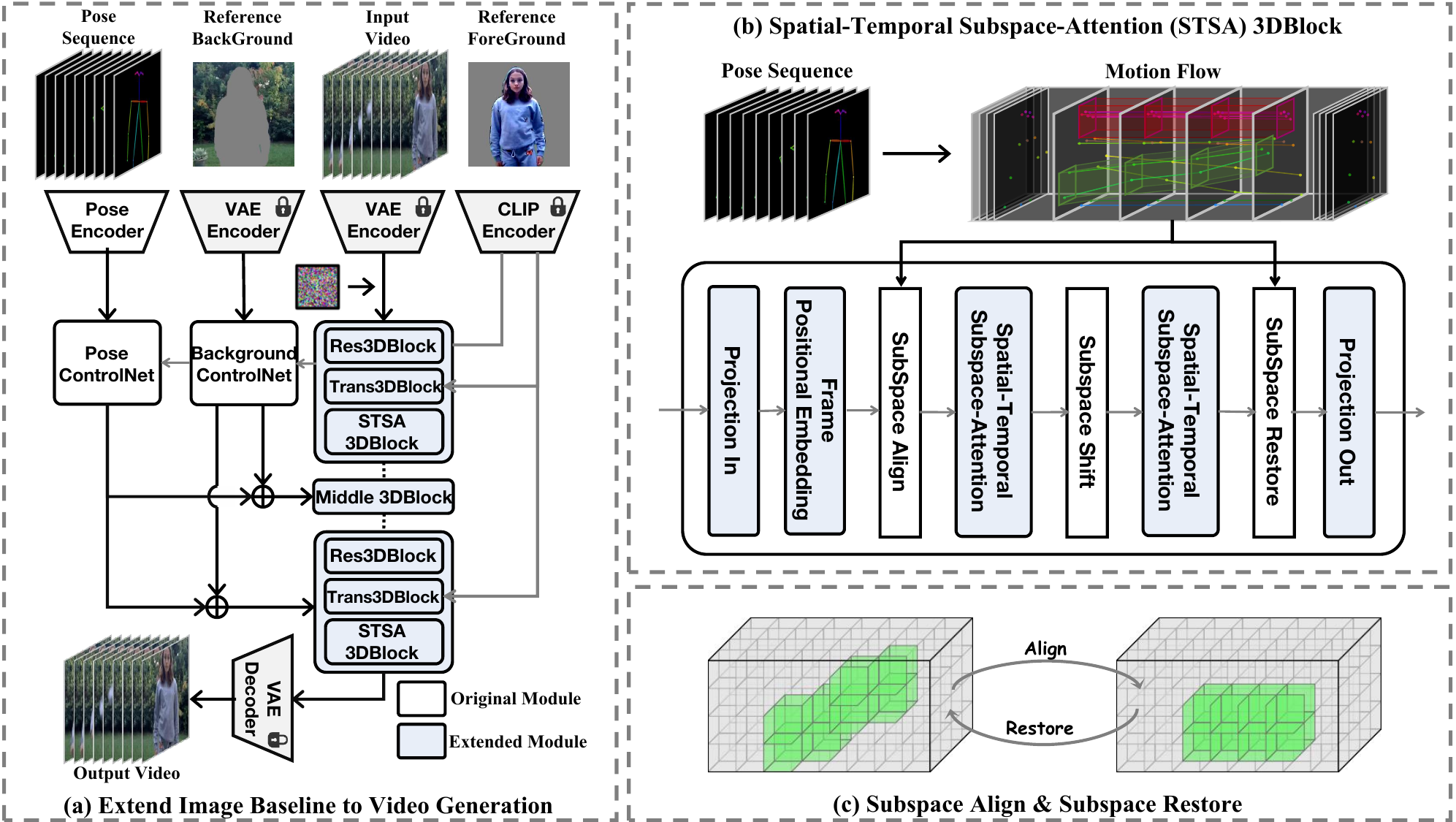}
 	\caption{Overview. (a) Based on DisCo, we extend the 2D U-Net to Pseudo 3D U-Net and introduce STSA blocks to generate videos. (b) We utilize coherent Motion Flow to guide Subspace Align \& Restore within the Spatial-Temporal Subspace-Attention mechanism, modeling each subspaces' spatiotemporal consistency and propagating consistency across subspaces through Subspace Shift operations.  (c) Example of feature maps through Subspace Align \& Restore, illustrating the alignment of content and position within subspaces.}
    \label{fig:overview}
\end{figure*}

\subsection{Spatial-Temporal Consistency Block.}

Certain techniques \cite{TuneAVideo, T2VZero, FateZero, ControlVideo} extend the self-attention module to incorporate information from multiple frames. For instance, Tune-A-Video computes the attention on the first and the previous frame. Text2Video-Zero introduces cross-frame attention by referencing the first frame. FateZero calculates the attention of each frame with the middle frame of the preceding frames. ControlVideo broadens self-attention by facilitating interaction across all frames. However, while these approaches enhance temporal consistency, they alter the inherent self-attention module, leading to a slight reduction in image generation quality. Additionally, some methods \cite{vdm, MakeAVideo} extend the 2D U-Net to a Pseudo 3D U-Net. This involves adjusting spatial and temporal dimensions to enable interactions across the same spatial positions at different temporal positions. Despite the effectiveness, these methods struggle with moving objects and are susceptible to ghosting and flickering artifacts. 

Furthermore, some methods leverage the direction of motion to enhance spatial-temporal consistency. For instance, Text2Video-Zero \cite{T2VZero} incorporates predetermined directions by introducing motion dynamics in latent codes. LFDM \cite{LFDM} generates motion videos by constructing an optical flow sequence based on prompt. In Human Dance Generation, DreamPose \cite{DreamPose} inputs five consecutive poses simultaneously to generate the target frame. In contrast to these methods, we aim to use motion flow extracted from pose sequences to enhance spatial-temporal consistency.
\section{Preliminaries}

\subsection{Latent Diffusion Models.}

LDM \cite{SD}, which operates within a latent space employing an auto-encoder and a U-Net, has shown significant performance in image generation. During the forward process, the latent input \(z_0\) is perturbed with Gaussian noise. The density of \(z_t\) conditioned on \(z_{t-1}\) can be expressed as follows:
\begin{equation}
q(z_t | z_{t-1}) = \mathcal{N}\left( z_t ; \sqrt{1 - \beta_t}{z_{t-1}, \beta_t}I \right),
\end{equation}
where \(\beta_t\) is the variance schedule at timestep \(t\). For denoising, the backward process can be formulated as:
\begin{equation}
p_\theta(z_{t-1} | z_t) = \mathcal{N}\left( z_{t-1} ; \mu_\theta(z_t, \tau, t), \Sigma_\theta(z_t, \tau, t) \right),
\end{equation}
where \(\tau\) symbolizes the textual prompt. The quantities \(\mu_\theta\) and \(\Sigma_\theta\) are computed by the denoising model \(\epsilon_{\theta}\).

\subsection{Diffusion-based Dance Generation.}
In the field of human dance generation, given a sequence of poses \( p = \{p_1, p_2, \ldots, p_T \} \), a human foreground \( f \), and a background \( g \), the aim is to synthesize a realistic video \( V = \{I_1, I_2, \ldots, I_T \} \) conditioned on \( f, g, p\). As discussed in related works, numerous commendable methods have been explored for this task. Among the diffusion-based approaches, DisCo \cite{DisCo} stands as the current SOTA method. Specifically, it incorporates disentangled control for \( f, g, p \), utilizing the Pose ControlNet \( \tau_\theta \) and Background ControlNet \( \mu_\theta \) to control \(p\) and \(g\), and employing the Cross Attention to control \(f\). The objective can be formulated as:
\begin{equation}
\small{L = \mathbb{E}_{f, p, g, \epsilon \sim \mathcal{N}(0,1)} \left[ \left\| \epsilon - \epsilon_\theta \left( z_t, t, f, \tau_\theta(p), \mu_\theta(g) \right) \right\|_2^2 \right]}.
\end{equation}

However, DisCo's frame-by-frame generation approach results in unsatisfactory spatiotemporal consistency. In our work, we choose this approach as a baseline, extend it into a video generation model, and then devise various techniques to enhance the spatiotemporal consistency, thereby enabling the generation of consistent human dance videos.
\section{Method}

\subsection{Overview.}
We introduce Dance-Your-Latents, our framework that guides latents to dance coherently following motion flow to generate consistent dance videos, as shown in Fig. \ref{fig:overview}. We aim to enhance the spatiotemporal consistency through spatial-temporal subspace attention guided by motion flow. 

We extend the original 2D U-Net to the Spatial-Temporal 3D U-Net by integrating spatiotemporal consistency modeling blocks to facilitate video generation, as illustrated in Fig. \ref{fig:overview}(a). Furthermore, as shown in Fig. \ref{fig:overview}(b), we extract motion flow from the pose sequence to guide the Subspace Align \& Restore operations, model and propagate the spatiotemporal consistency within each subspace through Subspace Attention and Subspace Shift, respectively. Additionally, Fig. \ref{fig:overview} (c) provides an illustration of the transformation of feature maps through Subspace Align \& Restore operations, highlighting the alignment within these subspaces.

\subsection{Extend Image to Video Generation.}
Firstly, We extend the frame-by-frame baseline model to a video generation model by changing the input pose and noise into sequences \(p = \{p_1, p_2, \ldots, p_T \}\) and \(z = \{z_1, z_2, \ldots, z_T \}\), respectively, with the aim of synthesizing a realistic video \( V = \{I_1, I_2, \ldots, I_T \} \) conditioned on \( f, g, p \). The disentangled control of \( f, g, p \) is maintained using ControlNet and Cross Attention. We transform the original 2D U-Net to a Spatial-Temporal 3D U-Net and introduce Spatial-Temporal Subspace-Attention Blocks to enhance the consistency of the generated videos.

\subsubsection{Spatial-Temporal 3D U-Net.}
As illustrated by the blue modules in Fig. \ref{fig:overview} (a), we extend the conventional 2D U-Net to Spatial-Temporal 3D U-Net following VDM \cite{vdm} by transforming the original ResBlock and TransBlock into Pseudo Res3DBlock and Trans3DBlock, respectivel, thereby enabling the model to efficiently process video inputs. Notably, unlike VDM, which employs modules such as TemporalConv and TemporalAttn, we introduce innovative Spatial-Temporal Subspace-Attention (STSA) 3DBlocks to model consistency more effectively. Furthermore, we incorporate frame positional embeddings to enable the network to distinguish the ordering of frames, and we synthesize each frame from the same Gaussian noise to simplify the complexity of modeling spatiotemporal consistency of different frames.

\subsubsection{Spatial-Temporal Subspace-Attention Block.}
It including two main components: Spatial-Temporal Subspace-Attention and Motion Flow Guided Subspace Align \& Restore. This structure is designed to decompose the global space into a combination of subspaces guided by coherent motion flow, model and propagate spatiotemporal consistency within and across these subspaces. 

We observe that most of the video's constituent elements moves within a small space, few elements traverse larger space but still much smaller than the global space, as shown in Fig. \ref{fig:motivation}. Based on our observations, we introduce the Spatial-Temporal Subspace-Attention. It decompose the global space into a combination of subspaces, utilizing Subspace Attention to model the spatiotemporal consistency within each subspace and Subspace Shift to propagate this spatiotemporal consistency across all subspaces.

Moreover, considering that the movement of each element is related to the motion, it is more reasonable to decompose the irregular subspace along with the motion flow. Consequently, we introduce Motion Flow Guided Subspace Align \& Restore. It extracts pure motion flow from the pose sequence, avoiding other interference from the original video, and guides the alignment and restoration of the content and position within each subspace. 

\subsection{Spatial-Temporal Subspace-Attention.}

\begin{figure}[t] 
	\includegraphics[width=8.4cm]{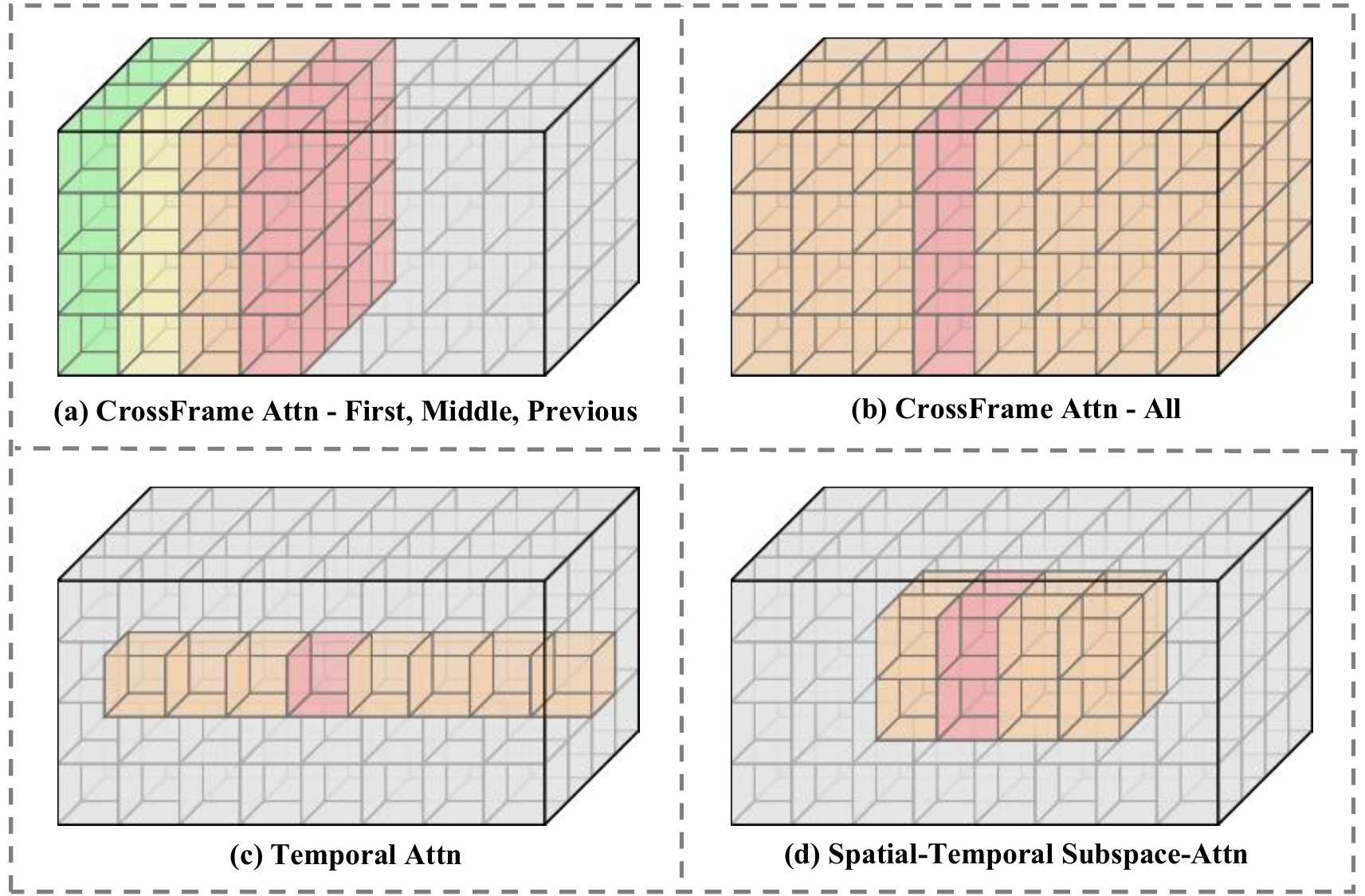}
 	\caption{Attention Methods. We designate patches of the key frame in red. (a) CrossFrame Attention, the colors green, yellow, and orange represent the first frame, middle frame, and previous frame, respectively. (b) CrossFrame Attention - All. (c) Temporal Attention. (d) Spatial-Temporal Subspace Attention.}
	\label{fig:attn}
\end{figure}

\subsubsection{Subspace Attention.}
Certain techniques \cite{TuneAVideo, T2VZero, FateZero, ControlVideo} extend the self-attention module to incorporate information from specific frame (e.g., first, middle and previous frame) or all frames, denoted as CrossFrameAttn and illustrated in Fig. \ref{fig:attn} (a) and (b). However, the benefits gained from long-distance attention are far outweighed by the significant increase in computational costs. Moreover, altering the fundamental self-attention module might result in a slight decrease in the quality of generated images.Some techniques \cite{vdm, MakeAVideo} incorporate a new TemporalAttn layer, which swaps between the spatial and temporal dimensions to facilitate interactions among patches at identical spatial positions but different temporal positions, as shown in Fig. \ref{fig:attn} (c). However, this method faces challenges in processing moving elements and sometimes produces visual artifacts such as ghosting. 

We present our Spatial-Temporal Subspace-Attention in Fig. \ref{fig:attn} (d). Distinct from other techniques, our attention mechanism operates on split, non-overlapping subspaces \(\mathcal{S} = \{\mathcal{S}_1, \mathcal{S}_2, \ldots, \mathcal{S}_N\} \) rather than the global space \(\mathcal{X}\). This subspace-based attention proves more effective, as it ensures each patch focuses on adjacent patches to avoid excessive spread of long-range attention. It can efficiently accommodate the motion variations while maintaining lower computational costs. The attention calculation within each subspace \(\mathcal{S}_k\) can be formulated as:
\begin{equation}
	Attention(\mathcal{S}_k) = Softmax(\frac{{Q_{\mathcal{S}_k} \cdot {K}_{\mathcal{S}_k}}^T}{\sqrt{d}})\cdot V_{\mathcal{S}_k},
\end{equation}
where \(Q_{\mathcal{S}_k}\), \(K_{\mathcal{S}_k}\), \(V_{\mathcal{S}_k}\) denote the query, key, value of \(\mathcal{S}_k\).

\subsubsection{Subspace Shift.}

Subspace Attention effectively models spatiotemporal consistency within individual non-overlapping subspaces, but it lacks connections across these subspaces, resulting in isolated spatiotemporal consistencies. To bridge these isolated subspaces and unify their spatiotemporal consistencies, we introduce a Subspace Shift operation on the feature map between two STSA modules. Assuming the subspace size is \(s=[s_f, s_h, s_w]\), inspired by the Swin-Transformer \cite{swin}, the shift size is set to be half the size of the subspace. We shift the subspaces \(\mathcal{S}=\{\mathcal{S}_1, \mathcal{S}_2, \ldots, \mathcal{S}_N\}\) by \((s_f/2, s_h/2, s_w/2)\) along the \((f, h, w)\) directions, resulting in the shifted subspaces \(\mathcal{S}'=\{\mathcal{S}'_1, \mathcal{S}'_2, \ldots, \mathcal{S}'_N\}\). Through the Subspace Shift operation, the spatiotemporal consistency is propagated across subspaces, ensuring overall spatiotemporal consistency throughout the global space.

\subsection{Motion Guided Subspace Align \& Restore.}

\subsubsection{Motion Flow.}

Optical flow describes the dense correspondences among voxels in adjacent video frames, and has been employed as a guide in some recent works on video translation \cite{videocontrolnet, tokenflow}. In the context of human dance generation, we aim that the generated video is guided solely by motion information, without being affected by the lighting, brightness, shape, or noise from the original video. Therefore, we devise a motion flow extraction method. Given two reference frames \(I_{i}, I_{j}\), we extract pose images \(p_{i}, p_{j}\), and then use RAFT \cite{raft} to estimate a dense pixel displacement field \(\mathcal{F}^{i\rightarrow j}\). We can obtain the coordinates of each key point \((i, x_i,y_i)\) in the \(i\)-th frame and their correspondences \((j,x_j,y_j)\) in the \(j\)-th frame. The flow \(\mathcal{F}=\{ \mathcal{F}_x, \mathcal{F}_y\}\) can be formulated as:
\begin{equation}
    (x_j,y_j) = (x_i + \mathcal{F}_{x}^{i\rightarrow j}(x_i, y_i), y_i + \mathcal{F}_{y}^{i\rightarrow j}(x_i, y_i)).
\end{equation}

Here, we choose DensePose's 3D mesh images \cite{densepose} over OpenPose's skeleton images \cite{openpose} for richer information. We downsample the dense motion flow to accommodate the resolution of the latent space in different modules of the U-Net.

\subsubsection{Subspace Align \& Restore.}
As shown in Fig. \ref{fig:overview} (c), with predetermined subspace size \(s=[s_f,s_h,s_w]\) and motion flow \(\mathcal{F}\), we construct an irregular subspace \(\mathcal{S}\). We introduce Subspace Align to convert \(\mathcal{S}\) to a regular subspace \(\tilde{\mathcal{S}}\) for efficient Subspace Attention, and Subspace recover the regular subspace to the original structure. By default, the central frame of each subspace is designated as the reference, serving to align patches from other frames.

For each patch located at \((k, x_k, y_k)\) in the \(k\)-th frame, we identify the beginning and ending frame index \(b\) and \(e\) of its associated subspace \(\mathcal{S}\), and then compute the reference frame index as \(r = \lfloor (b + e) / 2 \rfloor\). We employ \(\mathcal{F}^{i \rightarrow r}\) to compute the align target patch \((r,x_r,y_r)\) . It is noteworthy that a nearest neighbor calculation \(\psi\) is utilized for the discrete coordinates when the patch is divided. The comprehensive calculation is expressed as follows:
\begin{equation}
    (x_r,y_r) = (x_k + \mathcal{F}_{x}^{k\rightarrow r}(x_k, y_k), y_k + \mathcal{F}_{y}^{k\rightarrow r}(x_k, y_k)),
\end{equation}
\begin{equation}
    (x_r,y_r) = (\psi(x_r), \psi(y_r)).
\end{equation}

Subspace Align is achieved by moving the patch from \((k, x_k, y_k)\) to \((k, x_r, y_r)\), while Subspace Restore entails the reverse operation. Notably, it is unnecessary to perform Align and Restore operations discretely on each subspace \(\mathcal{S}\); instead, these operations can be simultaneously executed on the entire space \(\mathcal{X}\), yielding the aligned space \(\tilde{\mathcal{X}}\), and conducting Subspace Split to get all aligned subspaces \(\tilde{\mathcal{S}}\).
\begin{figure*}[t]
	\centering
	\includegraphics[width=0.85\linewidth]{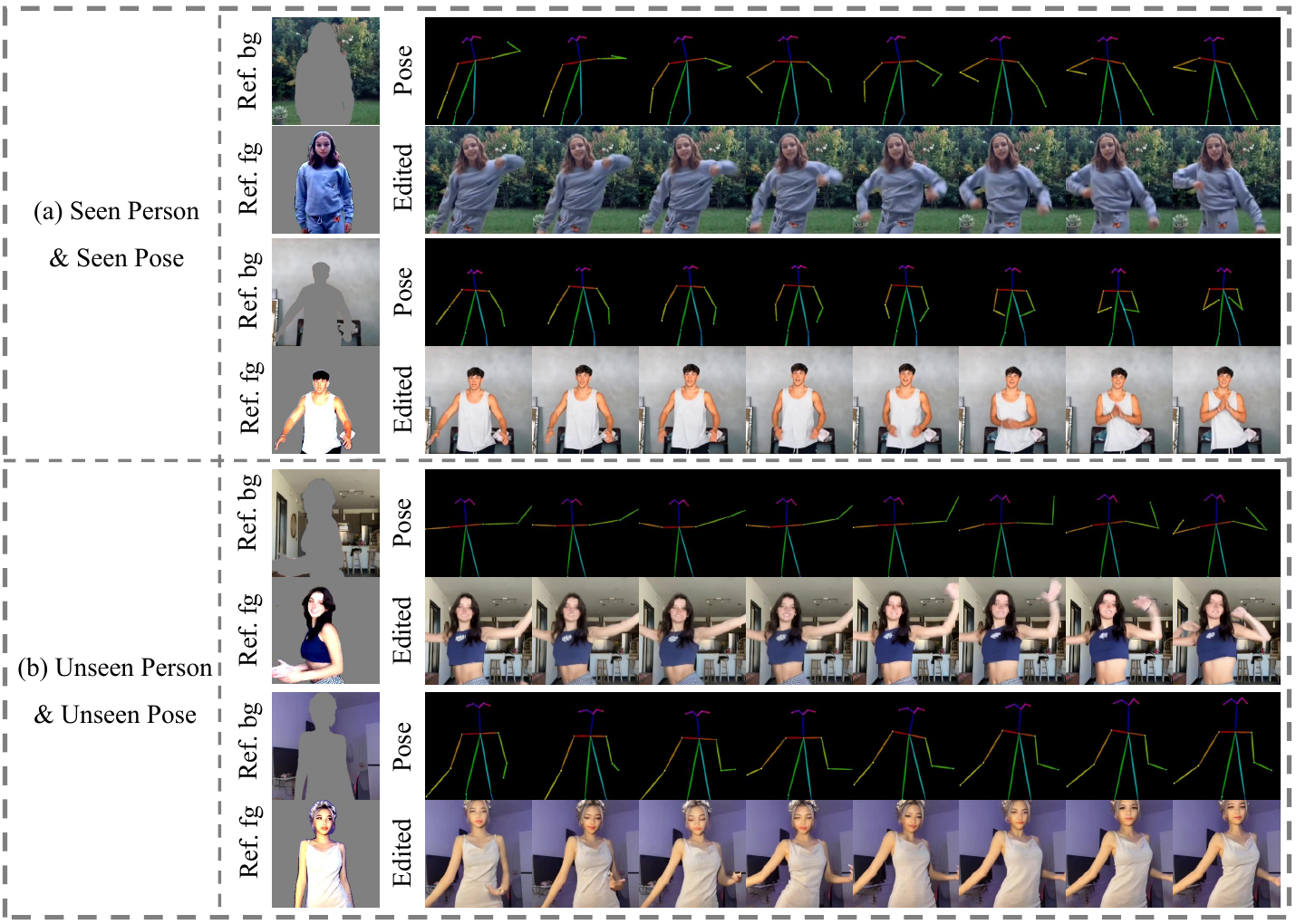}
	\caption{Applications. We opted for the simplest cases of (a) Seen Person and Seen Pose, as well as the most challenging cases of (b) Unseen Person and Unseen Pose, to showcase the range of capabilities of our model. Best viewed when zoomed-in.}
	\label{fig:applications}
\end{figure*}

\section{Experiments}

\subsection{Implementation Details.}

We utilized the TikTok dataset \cite{tiktok}, partitioned according to the DisCo split. This dataset contains 335 training videos and 10 testing videos. In our approach, we initially fixed the parameters of the baseline module and solely trained the STSA module for 50k steps, employing a learning rate of $2e^{-4}$. Following this, we fine-tuned all modules, excluding the Pose ControlNet and Background ControlNet, with a learning rate of $2e^{-5}$ for additional 20K steps. Unless otherwise mentioned, we maintain consistently used a frame size of 16 and a subspace size of $[4, 4, 4]$. All experiments were conducted on 8 NVIDIA V100 GPUs.

\subsection{Applications.} 

Our approach is characterized by its faithfulness, generalizability, and composability, demonstrating a robust capability in generating consistent human dance videos. We present four distinct scenarios, arranged in ascending order of complexity: (1) Seen Person \& Seen Pose, wherein the model has been exposed to both the persons and poses during training. (2) Seen Person \& Unseen Pose, where the model is familiar with the persons but encounters new poses. (3) Unseen Person \& Seen Pose, in which the model recognizes the poses but is introduced to new persons. (4) Unseen Person \& Unseen Pose: a scenario where both the persons and poses are unfamiliar to the model. To demonstrate the effectiveness of our Dance-Your-Latents in synthesizing consistent dance videos, we provide examples of both the simplest and most complex in Fig. \ref{fig:applications}. Additional videos can be found in the supplementary material.

\begin{table*}[h]
\centering
\resizebox{0.9\linewidth}{!}{
    \setlength{\tabcolsep}{12pt}{
    \renewcommand{\arraystretch}{1}

    \begin{tabular}{lccccccc}
    \toprule
    \multirow{2}[1]{*}{\textbf{Method}} & \multicolumn{5}{c}{\textbf{Image}} & \multicolumn{2}{c}{\textbf{Video}}\\
    \cmidrule(r){2-6}  \cmidrule(r){7-8} 
     & FID $\downarrow$ & SSIM $\uparrow$ & PSNR $\uparrow$ & LPIPS $\downarrow$ & L1 $\downarrow$ & FID-VID $\downarrow$ & FVD $\downarrow$\\
    \midrule
    DreamPose & 72.62 & 0.511 & 28.11 & 0.442 & 6.88E-04 & 53.36 & 671.50\\
    DisCo & 30.75 & 0.667 & 29.02 & \textbf{0.292} & 3.79E-04 & 24.70 & 562.01 \\
    \midrule
    Ours (w/. VD) & 32.13 & 0.665 & \textbf{29.11} & 0.294 & 3.77E-04 & 21.34 & 441.64 \\
    Ours (w/. VD, SA) & 30.88 & 0.667 & 29.08 & 0.293 & 3.78E-04 & 17.76 & 366.52 \\
    Ours (w/. VD, SA, MF) & \textbf{29.30} & \textbf{0.671} & 29.10 & \textbf{0.292} & \textbf{3.74E-04} & \textbf{15.83} & \textbf{334.81} \\
    \bottomrule
\end{tabular}
}}
\caption{Comparisons. ``VD'' represents Video Dance Diffusion extended by our approach. ``SA'' stands for Spatial-Temporal Subspace Attention. ``MF'' denotes Motion Flow guided Space Align $\&$ Restore. $\downarrow$ indicates the lower the better, and vice versa.}
\label{table:sota}
\end{table*}

\begin{figure*}[h]
	\centering
	\includegraphics[width=1.0\linewidth]{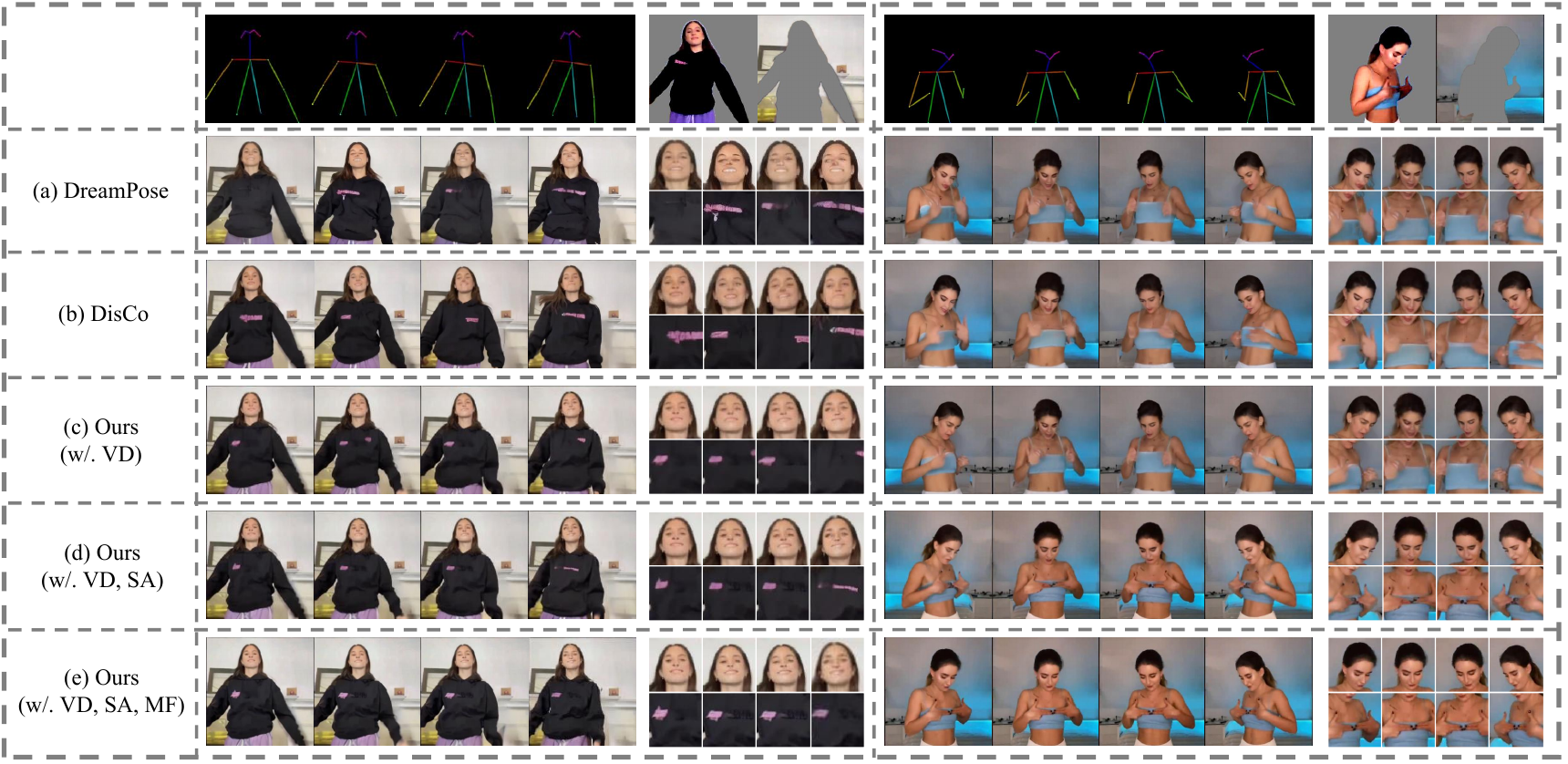}
 	\caption{Comparisons. (e) vs. (a,b): more consistent and detailed faces, stabilized logo positions and shapes, and coherent hands movements; (c) vs. (b): enhanced consistency in color tone and facial details; (d) vs. (c): logo moves coherently, no elements jitter because of Subspace Attention; (e) vs. (d): stabilized logo positions and shapes, hand movements are more coherent and enriched with detail.  }
	\label{fig:comparision}
\end{figure*}

\subsection{Comparisons.}

\subsubsection{Quantitative Results.} 

In line with DisCo\cite{DisCo}, we employ FID \cite{FID}, SSIM \cite{SSIM}, LPIPS \cite{LPIPS}, and L1 to evaluate image quality, and utilize FID-VID \cite{FIDVID} and FVD \cite{FVD} for consecutive 16 frames to evaluate video quality. We report the quantitative results in Tab. \ref{table:sota}. Compared with state-of-the-art methods like DreamPose \cite{DreamPose} and DisCo \cite{DisCo}, our approach significantly outperforms others in both image and video quality. Specifically, by extending the baseline model to the video generation model (VD), where all frames start denoising from the same latent space and are guided by the same human foreground and background, the consistency notably improves while the image quality fluctuates within acceptable limits. Moreover, our design of the novel spatial-temporal subspace-attention (SA) and motion flow (ML) guidance not only slightly enhances image generation quality but also substantially improves video consistency. 

\subsubsection{Qualitative Results.}
We present a comparison of the videos generated by our Dance-Your-Latents and previous state-of-the-art methods, including the three mentioned improvements, as illustrated in Fig. \ref{fig:comparision}. Comparing (e) with (a) and (b), it is evident that our approach generates videos of superior quality compared to DreamPose and DisCo, as manifested in the more consistent and detailed faces, stabilized logo positions and shapes, and coherent hands movements depicted in the figure.

Specifically, comparing (b) with (c), our method exhibits enhanced consistency in color tone and facial details, attributable to the fact that all frames initiate the denoising process from the same latent space and are guided by the same human foreground and background. In the comparison between (d) and (c), we can observe noticeable logo position jitter between adjacent frames in (c), whereas in (d), the logo moves coherently towards the rightward direction. Although this movement pattern is not reasonable in real-world scenarios, the consistency is markedly improved. Additionally, upon comparing (e) with (b), it is evident that both the position and shapes of the logo in (e) remain stable, and the hand movements are more coherent and enriched with detail. This consistency is a result of our Motion Flow-guided Subspace Align \& Restore, which ensures both content and positional alignment. The experiments demonstrate the superiority of our approach.

\subsection{Ablations.}

\subsubsection{Effect of Subspace Attention}

In Tab. \ref{table:ablation_attention}, we conducted an ablation study to evaluate the effect of Subspace Attention guided by Motion Flow. We observed that CrossFrame Attn - First, Middle, Previous and All display a low FID at the image level. This phenomenon is attributed to the modification of the original self-attention structure, which results in a degradation of image quality. Simultaneously, these methods showcase suboptimal metrics at the video level, mainly due to their reliance on long-range dependencies, which in turn result in attention dispersion and compromised video quality. In contrast, our method significantly outperforms the CrossFrame Attn methods by preserving the original self-attention module and introduce the STSA block to model the spatiotemporal consistency. Furthermore, our approach demonstrates superiority over TemporalAttn, notably in video level metrics. This is attributed to Temporal Attn's inability to model extensive motion coherence, while we can align these subspaces guided by motion flow, thereby facilitating efficient attention calculations in irregular subspaces.

\subsubsection{Effect of Subspace Size}

Furthermore, we investigated the influence of Subspace Size in Tab. \ref{table:ablation_subspace}. We conducted experiments with three different subspace sizes: $[4,2,2]$, $[4,4,4]$, and $[8,4,4]$. The results indicate that the $[4,4,4]$ subspace size achieved the best performance. This outcome can be ascribed to the consideration that a temporal size of $8$ might extend the attention span excessively, whereas a spatial size of $2\times2$ is too limited to effectively capture larger movements of patches.

\begin{table}[t]
\centering
\resizebox{\linewidth}{!}{
    \setlength{\tabcolsep}{8pt}{
    \renewcommand{\arraystretch}{1}
    
    \begin{tabular}{lccc}
        \toprule
        \multirow{2}[1]{*}{\textbf{Attention}} & \multicolumn{1}{c}{\textbf{Image}} & \multicolumn{2}{c}{\textbf{Video}}\\
        \cmidrule(r){2-2}  \cmidrule(r){3-4} 
         & FID $\downarrow$ & FID-VID $\downarrow$ & FVD $\downarrow$\\
        \midrule
        CrossFrame Attn - First& 34.82 & 21.68 & 451.25   \\
        CrossFrame Attn - Middle& 32.46 & 21.25 & 434.70   \\
        CrossFrame Attn - Previous& 30.95 & 18.89 & 396.27   \\
        CrossFrame Attn - All& 35.73 & 19.04 & 402.72  \\
        \midrule
        Temporal Attn & 31.25 & 19.13 & 402.34  \\
        Subspace Attn (Ours) & \textbf{30.88} & \textbf{17.76} & \textbf{366.52} \\
        \bottomrule
    \end{tabular}}
}
\caption{Ablation of Subspace Attention.}
\label{table:ablation_attention}
\end{table}

\begin{table}[t]
\centering
\resizebox{\linewidth}{!}{
    \setlength{\tabcolsep}{14pt}
    {
    \begin{tabular}{lccc}
        \toprule
        \multirow{2}[1]{*}{\textbf{Subspace Size}} & \multicolumn{1}{c}{\textbf{Image}} & \multicolumn{2}{c}{\textbf{Video}}\\
        \cmidrule(r){2-2}  \cmidrule(r){3-4} 
         & FID $\downarrow$ & FID-VID $\downarrow$ & FVD $\downarrow$\\
        \midrule
        $[4,2,2]$ & 30.90 & 18.02 & 386.96   \\
        $[4,4,4]$ & \textbf{30.88} & 17.76 & \textbf{366.52} \\
        $[8,4,4]$ & 31.21 & \textbf{17.68} & 368.15   \\
        \bottomrule
    \end{tabular}}
}
\caption{Ablation of Subspace Size.}
\label{table:ablation_subspace}
\end{table}
\section{Conclusion}

In this paper, we propose Dance-Your-Latents, an approach aims to guide latents dance coherently following motion flow to generate consistent dance videos. Specifically, we first extend the 2D U-Net to Pseudo 3D U-Net to facilitate video generation. Then, considering each constituent element moves within a confined space, we introduce the spatial-temporal subspace attention blocks. These blocks decompose the the global space into a combination of regular subspaces and efficiently models and propagate the spatiotemporal consistency within and across these subspaces. Furthermore, we extract motion flows from pose sequences and utilize them to align content and position within each subspace, guiding attention to regions along the motion flow. Experiments demonstrate our efficiency in achieving superior spatiotemporal consistency of generated videos.

{
    \small
    \bibliographystyle{ieeenat_fullname}
    \bibliography{main}
}


\end{document}